\documentclass[journal]{IEEEtran}
\IEEEoverridecommandlockouts
\usepackage{cite}
\usepackage{amsmath,amssymb,amsfonts}
\usepackage{algorithmic}
\usepackage{graphicx}
\usepackage{textcomp}
\usepackage{xcolor}
\usepackage{comment}
\usepackage{hyperref}
\usepackage[font=small,labelfont=bf]{caption}
\usepackage{amsmath}
\usepackage{algorithm}
\usepackage{amsmath}
\usepackage{amssymb}
\usepackage{placeins}
\usepackage{subfiles}
\usepackage{multirow}
\usepackage{multicol}
\usepackage{enumitem}
\usepackage{subfigure}
\def\BibTeX{{\rm B\kern-.05em{\sc i\kern-.025em b}\kern-.08em
    T\kern-.1667em\lower.7ex\hbox{E}\kern-.125emX}}

\renewcommand{\footnoterule}{%
  \kern -3pt
  \hrule width 0.4\columnwidth height 0.5pt
  \kern 2pt
}

\begin{document}
\title{Probing Image Compression\\ For Class-Incremental Learning}

\author{\IEEEauthorblockN{Justin Yang, Zhihao Duan, Andrew Peng, Yuning Huang, Jiangpeng He\(^{\dagger}\), Fengqing Zhu}\\
\IEEEauthorblockA{Elmore School of Electrical and Computer Engineering, Purdue University, West Lafayette, Indiana, USA}
}
\maketitle

\renewcommand{\thefootnote}{\fnsymbol{footnote}} 
\footnotetext[2]{Corresponding author}

\begin{abstract}
Image compression emerges as a pivotal tool in the efficient handling and transmission of digital images. Its ability to substantially reduce file size not only facilitates enhanced data storage capacity but also potentially brings advantages to the development of continual machine learning (ML) systems, which learn new knowledge incrementally from sequential data. Continual ML systems often rely on storing representative samples, also known as exemplars, within a limited memory constraint to maintain the performance on previously learned data. These methods are known as memory replay-based algorithms and have proven effective at mitigating the detrimental effects of catastrophic forgetting. Nonetheless, the limited memory buffer size often falls short of adequately representing the entire data distribution. In this paper, we explore the use of image compression as a strategy to enhance the buffer's capacity, thereby increasing exemplar diversity. However, directly using compressed exemplars introduces domain shift during continual ML, marked by a discrepancy between compressed training data and uncompressed testing data. Additionally, it is essential to determine the appropriate compression algorithm and select the most effective rate for continual ML systems to balance the trade-off between exemplar quality and quantity. To this end, we introduce a new framework to incorporate image compression for continual ML including a pre-processing data compression step and an efficient compression rate/algorithm selection method. We conduct extensive experiments on CIFAR-100 and ImageNet datasets and show that our method significantly improves image classification accuracy in continual ML settings.


\end{abstract}
\section{Introduction}
\label{sec:Introduction}
The core aim of continual machine learning is to emulate human-like learning in AI systems, enabling ML systems to acquire and build upon knowledge from an ongoing flow of data throughout their operational lifespan. The major challenge is catastrophic forgetting, a phenomenon where neural network models tend to lose previously acquired information when exposed to new data. In the realm of continual learning, class-incremental learning (CIL) stands out as particularly challenging yet representative of real-world scenarios where the CIL model is expected to classify for all classes encountered so far during the inference. In this paper, we focus on CIL, utilizing memory replay, an effective method against catastrophic forgetting. Memory replay involves retaining selected exemplars from previous classes within a defined memory budget, aiding in balancing new and old data during model training~\cite{PODNet, iCaRL, mnemonics, Liu2021RMM}. However, the performance of memory-replay-based methods greatly relies on the memory budget, which is a significant constraint in CIL. Therefore, we aim to address this issue in this work through the use of image compression to store more exemplars without increasing the memory budget.


Image compression allows a larger volume of previously seen class data in the memory buffer, promoting a more balanced and diverse training set. This is especially beneficial when utilizing low-bitrate compression, which significantly increases data storage efficiency compared to standard compression methods. However, directly applying compression to exemplars can lead to degradation in classification accuracy, primarily due to the domain shift between the source's compressed data and the target's original testing data. A previous study~\cite{janeiro2023visual} working on vision tasks with compression data had suggested fine-tuning models with target domain data to mitigate this issue, but this is not feasible in continual learning scenarios where original uncompressed data becomes unavailable in later training stages. This highlights the need to address performance degradation to fully benefit from leveraging low-bitrate compression.


In this work, we first propose to use data rate to determine the equivalent exemplar size for CIL methods for utilizing compressed exemplars. We then address the domain shift issue in CIL setups when utilizing compression by employing a pre-processing step to align the data characteristics between the training and testing phases. We investigate various image compression methods, including traditional~\cite{wallace1992jpeg} and learned~\cite{ding2021advances} ones, on both uncompressed datasets and compressed datasets. Additionally, we introduce an efficient framework utilizing only first-step data to select the suitable compression rate and algorithm to balance the trade-off between exemplar quality and quantity, and conduct extensive experiments to evaluate existing CIL methods with compressed exemplars. 
\section{Problem Formulation and Methods}
\label{sec:method}
In this section, we first formulate the class-incremental learning problem and introduce the calculation of equivalent memory using bitrates.
Subsequently, we delve into the challenges posed by the domain shift problem arising from the compressed exemplars and illustrate our method to mitigate this issue. 
Finally, we leverage forgetting measure~\cite{Chaudhry_2018_ECCV} with initial step data to identify compression rates to balance the trade-off between exemplar quality and quantity, and propose to use feature MSE to determine the best compression method for exemplar compression configuration. 

\subsection{Problem Formulation and Equivalent Memory}

\label{subsec:Equal_Memo}
Class-incremental learning (CIL) for image classification involves learning a sequence of $N$ training tasks defined as \{$T^0, T^1, ....T^{N-1}$\} without overlapping classes. Each task is denoted as $T^c = \{\left(\mathbf{x}^c_j, {y}^c_j\right)\}^{M}_{j=1}$, where $c<N$ and contains a total of $M$ training data. $x^c_j$ is the image data, and $y^c_j$ is the label for the $c^{th}$ task. After learning each task, the CIL model is expected to classify all classes seen so far. 

For a given task $T^c$, we define the exemplar set as an extra collection of training data from previous tasks, denoted as $\mathcal{E}=\left\{\left(\mathbf{x}^p_j, y^p_j\right)\right\}_{j=1}^{B}$. Here $p<c$, and $B$ represents the memory constraint, which determines the maximum number of original images the buffer can store. By utilizing the exemplar set, the model can incorporate $\mathcal{E} \cup \mathcal{T}^c$ training data during the update process for each task. The selection of the exemplar set for each task is determined using herding~\cite{herding}, which selects exemplar based on the distance between the feature of the samples and the average feature of the class.

In image compression, the data rate is typically quantified by bits per pixel (bpp), calculated by dividing the total size of the compressed images (in bits) by the total number of pixels in the image. Determining equivalent memory settings of exemplars using data rate measured in bpp under different compression methods is important for fair comparison. This is achieved by establishing a common data rate for different compression methods, thereby ensuring that each exemplar setting consumes an equal amount of data. The intent behind this process is to standardize memory utilization across different compression settings, ensuring fair representation and memory allocation in the continual learning context, regardless of the original size or the resolution of the input image. 

\begin{figure}[h]
    \centering
    \subfigure[]{\includegraphics[width=0.24\textwidth]{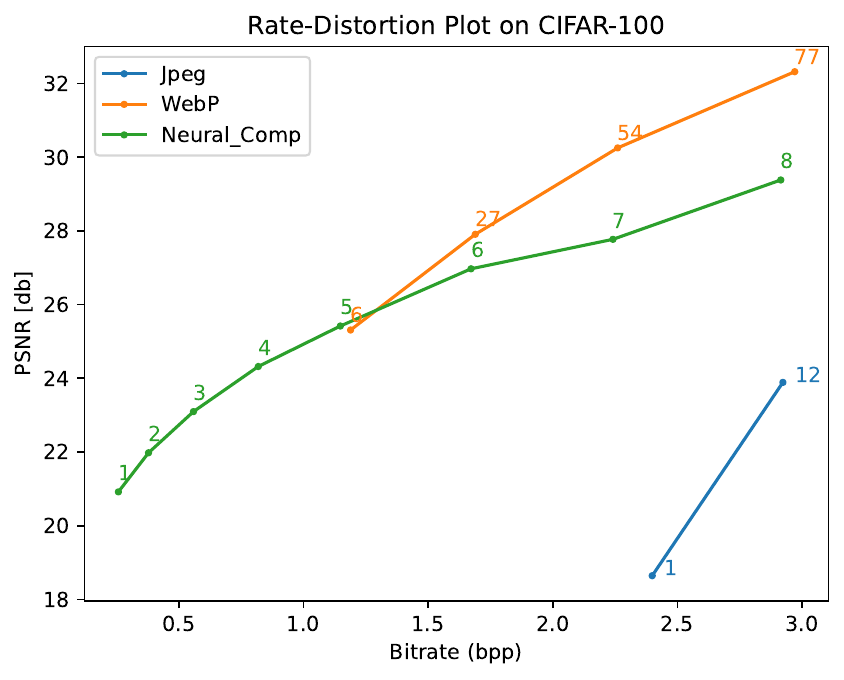}} 
    \subfigure[]{\includegraphics[width=0.24\textwidth]{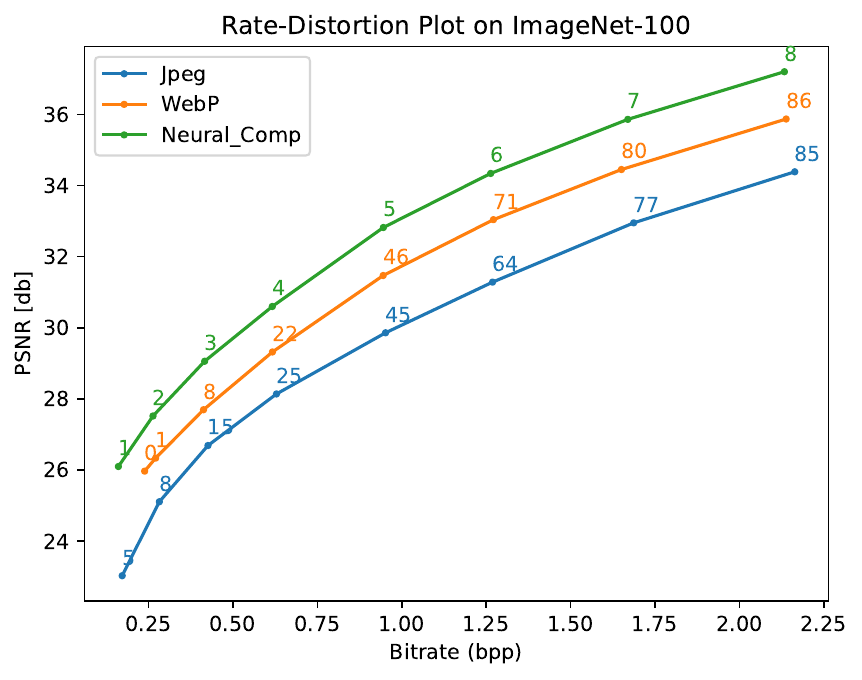}}
\vspace{-0.3cm}
\caption{Comparative analysis of different compression methods for (a) CIFAR-100 and (b) ImageNet-100 datasets. The x-axis represents data rate (bpp), while the y-axis denotes the Peak Signal-to-Noise Ratio (PSNR). Numbers on the graph highlight the compression level.}\vspace{-0.3cm}
\label{fig:dataset_rd}
\end{figure}

With the introduction of compression, the exemplar set for task $T^c$ is now represented as $\mathcal{E}_{comp}=\left\{\left(\mathbf{x}^p_j, y^p_j\right)\right\}_{j=1}^{B'}$, where  $p<c$. The value of $B'$ is determined by Equation \ref{eq:Buffersize}.
 \begin{equation}
    \centering
    \begin{aligned}
        \label{eq:Buffersize}
        B'=B\times (\tfrac{{bpp}_{ori}}{{bpp}_{comp}})
    \end{aligned}
\end{equation}
Since ${bpp}_{ori}$ is larger than ${bpp}_{comp}$, the updated memory constraint $B'$ allows more images to be included as exemplars. 
Here ${bpp}_{ori}$ and ${bpp}_{comp}$ are the average bpp for the original dataset and the compressed dataset, respectively. For a fair comparison, we focus on the date rate region where all compression methods overlap and calculate the number of exemplars under equivalent memory using the bpp ratio between compressed data and original data. Figure \ref{fig:dataset_rd} shows the initial set of all of the compression methods and levels of CIFAR-100 and ImageNet-100 used in our experiments, where we later determine the suitable rate\& algorithm combination.


\begin{figure}[h]
    \centering
    \subfigure[]{\includegraphics[width=0.24\textwidth]{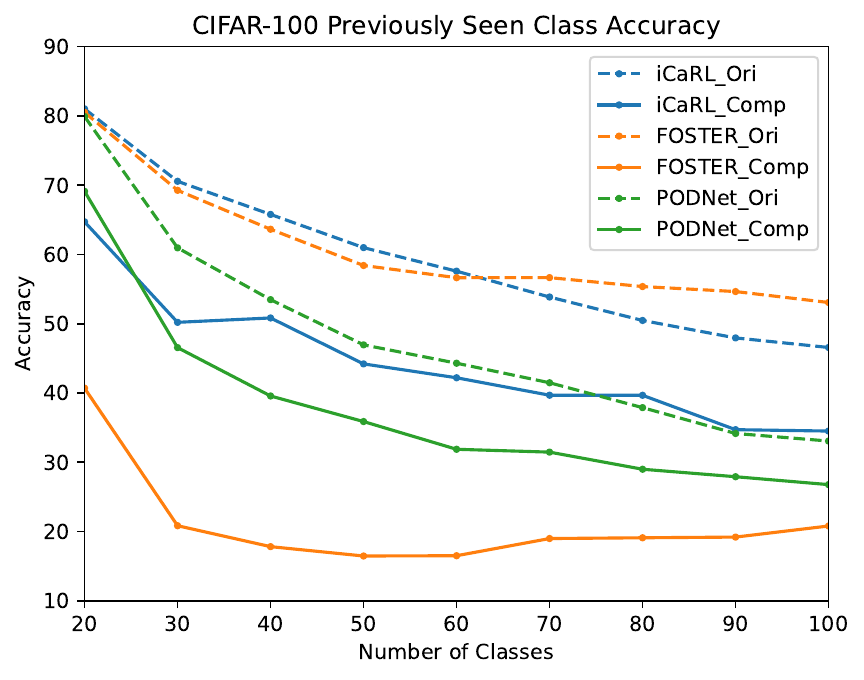}} 
    \subfigure[]{\includegraphics[width=0.24\textwidth]{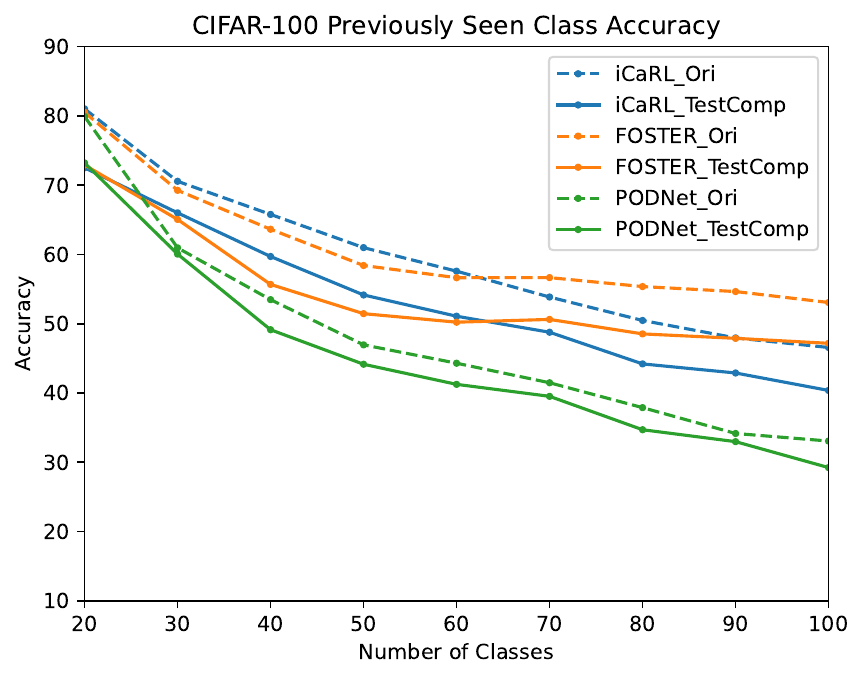}}
\vspace{-0.3cm}
\caption{Previously seen class accuracy with and without compression pre-processing. The testing set is not compressed for (a) and showed significant performance degradation compared with (b) where the testing set is compressed during pre-processing. Note that the model is trained on compressed exemplars and original new class data.} \vspace{-0.3cm}
  \label{fig:domainShift}
\end{figure}

\subsection{Domain Shift Issue}
Recent works~\cite{CIM_CIL,Wang2022MemoryRW} mainly focus on compressing the exemplars for CIL. However, directly using compressed images as exemplars would impose domain shift problems in CIL. Similar issues have been noted in a recent study focused on various vision tasks \cite{janeiro2023visual}. Low data rate compression significantly impacts the recognition capabilities of neural networks, primarily because of the discrepancy between the compressed and original data. This discrepancy also happens between compressed exemplars and the uncompressed testing set in the CIL context. Such shift appears to precipitate a substantial decrease in accuracy for previously learned classes at each stage, in contrast to the results obtained when the compression settings for exemplars and the testing set are consistent, as shown in Figure~\ref{fig:domainShift}(a), where we conduct experiments on several CIL baselines~\cite{iCaRL, Foster, PODNet} with exemplar compression. A potential solution to address this issue is to compress the previously seen classes data during inference. However, since task identifiers are not provided in the CIL, recognizing the data that belongs to learned classes is a challenging task. Previous works~\cite{CIM_CIL,Wang2022MemoryRW} deal with this by using compression methods with high data rates, but this limits the benefit of compression. Hence we more focus on low data rate compression. To tackle the performance degradation resulting from domain shift, we propose to perform compression as a pre-processing step across the entire dataset, including both the training and testing phases. This simple strategy is designed to mitigate the potential domain shift that would cause severe performance degradation during the testing stage. 
\begin{figure}[h]
    \centering
    \subfigure[]{\includegraphics[width=0.24\textwidth]{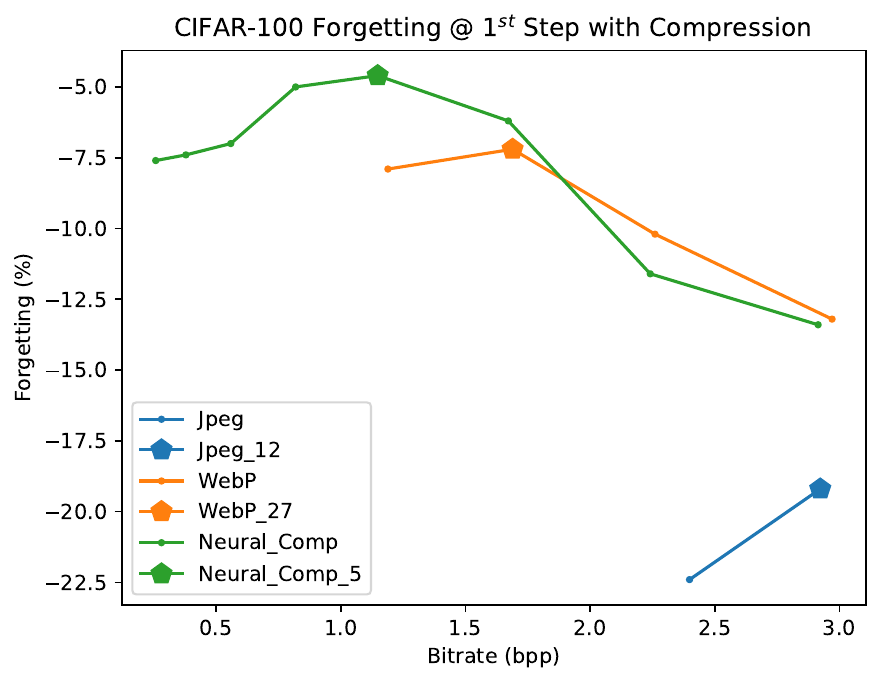}} 
    \subfigure[]{\includegraphics[width=0.24\textwidth]{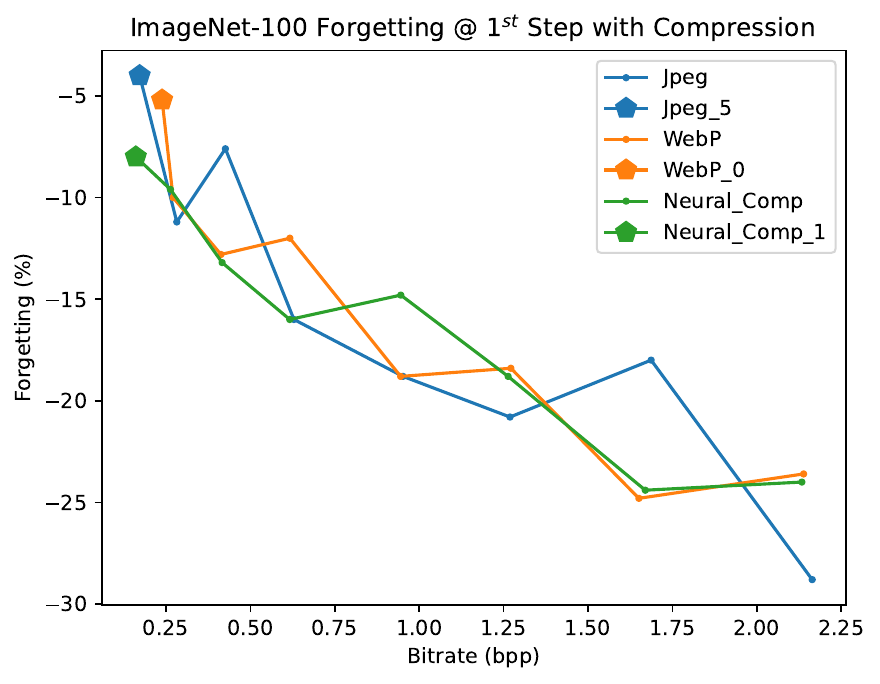}} 
\vspace{-0.3cm}
\caption{Forgetting results for each compressor under different rate settings. Note that the selected rate settings used in the latter experiments are shown in \textbf{pentagon}. }  \vspace{-0.6cm} \label{fig:forgetting}
\end{figure}

\subsection{Selection of Compression Rate and Algorithms}
\label{subsec:selection}
\subsubsection{Rate selection}
\label{subsec:suitableRes}
In this section, we sought to determine the rate setting for representative compression methods including JPEG, WebP, and neural compression, and then determine the suitable compression method in the later section. For neural compressors, we choose the Factorized model \cite{balle18hyperprior} for CIFAR-100 and the Scale Hyperprior model \cite{balle18hyperprior, minnen2018joint} for ImageNet-100. The most straightforward approach to determining compression rate would be to conduct a grid search across different compression rates. However, conducting multiple experiments covering all compression rates, \textit{e.g.} 100 different compression rates for JPEG and WebP, would significantly increase computational costs. This underscores the importance of selecting the appropriate rate, especially for the practical deployment of continual learning. Also, in real-world scenarios, a continual ML system is trained on sequential data. It neither stores all the data nor performs multiple runs using various compression methods and settings for hyperparameter adjustment due to storage limitations.

To tackle this issue, we propose to determine the rate for each compression method using only data from $T^0$. To be more specific, we partition $T^0$ into two equal halves, namely $T^0_1$ and $T^0_2$, where $T^0_1 = \{\left(\mathbf{x}^0_j, {y}^0_j\right)\}^{M/2}_{j=1}$ and $T^0_2 = \{\left(\mathbf{x}^0_j, {y}^0_j\right)\}^{M}_{j=M/2+1}$. We train a simple 2-step incremental learning model that updates the parameters on $T^0_2$ with exemplars, which are selected based on herding~\cite{herding}, after training on $T^0_1$ to assess the forgetting measure~\cite{Chaudhry_2018_ECCV}, which is the accuracy on $T^0_1$ using the first step model, subtracted by the accuracy on $T^0_1$ using the second step model, under different rate settings. It is important to note that the rate determined in this way aligns with the results training a more complex incremental learning model. We select the rate setting that yields the least forgetting for subsequent experiments. The rates selected for each compression algorithm for CIFAR-100 and ImageNet-100 are presented as pentagons in Figure~\ref{fig:forgetting}. For CIFAR-100, JPEG12, WebP27, and Factorized model with quality=5 are used. In the case of ImageNet-100, JPEG5, WebP0, and Scale Hyperprior model with quality=1 were deployed.

\begin{table}[h]
    \centering
    \caption{Compression algorithm selection on CIFAR-100. \textbf{Bold} indicates the final selected compression algorithm for experiments.} 
    \footnotesize
    \begin{tabular}{|c|c|c|c|c|} 
        \hline Method& Bitrate& PSNR (db) & $F_{MSE}$& Final Acc.\\ 
        \hline JPEG& 2.923 & 23.89 & 0.177 & 39.79\%\\
        \hline WebP & 1.689 & 27.91 & 0.113& 53.93\%\\
        \hline \textbf{Neural Compressor} & \textbf{1.148} & \textbf{25.42} & \textbf{0.084}& \textbf{55.08\%} \\
        \hline
    \end{tabular}
    \label{tab:metric_cifar100}\vspace{-0.3cm}
\end{table}

\begin{table}[h]
    \centering
    \caption{Compression algorithm selection on ImageNet-100. \textbf{Bold} indicates the final selected compression algorithm for experiments.}
    \footnotesize
    \begin{tabular}{|c|c|c|c|c|} 
        \hline Method& Bitrate& PSNR (db) & $F_{MSE}$& Final Acc.\\ 
        \hline JPEG& 0.172 & 23.03 & 0.0659 & 70.2\%\\
        \hline WebP & 0.239 & 25.97 & 0.0569& 68.66\%\\
        \hline \textbf{Neural Compressor} & \textbf{0.161} & \textbf{26.1} & \textbf{0.0531}& \textbf{71.22\%} \\
        \hline
    \end{tabular}
    \label{tab:metric_imageNet100}
\end{table}

\subsubsection{Algorithms selection}
\label{subsec:algoSelect}
After pinpointing the suitable compression rates for each algorithm, it is also crucial to determine the most suitable compression method for a given dataset. While Peak Signal-to-Noise Ratio (PSNR) serves as a common metric for assessing image quality, it may not always align with classification accuracy. For instance, as shown in Table~\ref{tab:metric_cifar100} and Table~\ref{tab:metric_imageNet100}, WebP exhibits relatively high PSNR values (27.91 db and 25.97 db), indicating less distortion. However, less distortion does not correlate to better classification accuracy (53.93\% and 68.66\%). The accuracy reported here is the last-step classification accuracy of iCaRL~\cite{iCaRL}, a classic and simple CIL method, but the observed trend generally applies to other CIL methods as well. Due to the inconsistency between existing metrics and last-step accuracy, it would be valuable to design a metric that better aligns with classification accuracy. The classification network can be separated into two parts: a feature backbone network $\phi$, which is a ResNet32~\cite{resnet} for CIFAR-100 and ResNet18 for ImageNet-100, and a simple fully-connected layer $w$ as the classifier. The visual features extracted by the ResNet backbone are pivotal in the classification process, as they provide the foundational information upon which $w$ makes its predictions. Hence, the distortion between the original and compressed image features might provide insights into potential classification performance. We propose to use the feature difference to assess the distortion from the neural network's perspective. Feature difference is defined as Equation \ref{eq:FMSE}, 
 \begin{equation}
    \centering
    \begin{aligned}
        \label{eq:FMSE}
        F_{MSE} = \frac{1}{M} \sum_{i=1}^{M} (\phi(x_i) - \phi(x^{'}_{i}))^2
    \end{aligned}
\end{equation}
where $\phi(x_i)$ is the output feature vector given original image data $x_i$ and $\phi(x^{'}_{i})$ is the output feature vector given compressed image data $x^{'}_{i}$. In both Table~\ref{tab:metric_cifar100} and~\ref{tab:metric_imageNet100}, the results show that the compression algorithm with the lowest $F_{MSE}$, which is Neural Compressor in both datasets, leads to a better accuracy. Hence we selected Factorized model with quality=5 for CIFAR-100 and Scale Hyperprior model with quality=1 for ImageNet-100 as the suitable compression algorithm.

\vspace{-0.3cm}
\section{Experiments and Discussion}
\label{sec:experiment}

\begin{table*}[h]
    \centering
    \begin{tabular}{|c|c|c|c|c|c|c|c|c|c|c|c|c|}
    \hline \multirow{3}{*}{$\begin{array}{c}\text { Datasets } \\
    \text {Memory Buffer Size} \\
    \text { Accuracy (\%) } \\
    \end{array}$} & \multicolumn{4}{|c|}{ CIFAR-100 } & \multicolumn{4}{|c|}{ ImageNet-100 } \\
    \hline & \multicolumn{2}{|c|}{2.93MB} & \multicolumn{2}{|c|}{5.86MB} &\multicolumn{2}{|c|}{143.5MB} & \multicolumn{2}{|c|}{287MB} \\
    \hline & Avg & Last & Avg & Last & Avg & Last & Avg & Last \\
    \hline iCaRL~\cite{iCaRL} \textit{w/o} compression & 56.67 & 38.52 & 61.0 & 45.68 & 60.45 & 43.26 & 66.26 & 51.96 \\
    \hline iCaRL \textit{w/} JPEG & 51.68 & 37.54 & 53.14 & 39.79 & 73.34 & 65.9 & 75.48 & 70.2 \\
    \hline iCaRL \textit{w/} WebP & 63.34 & 50.61 & 65.24 & 53.93 & 71.4 & 63.1 & 74.46 & 68.66 \\
    \hline iCaRL \textit{w/} Neural Compressor & \textbf{63.6} & \textbf{52.16} & \textbf{64.92} & \textbf{55.08} & \textbf{73.87} & \textbf{66.32} & \textbf{76.44} & \textbf{71.22}\\ \hline
    \hline WA~\cite{wa2020} \textit{w/o} compression & 60.24 & 45.15 & 62.66 & 48.46 & 63.1 & 46.32 & 66.65 & 52.94 \\
    \hline WA \textit{w/} JPEG & 51.36 & 38.17 & 52.9 & 41.41 & 72.41 & 65.82 & 74.94 & 70.84 \\
    \hline WA \textit{w/} WebP & 62.2 & 51.24 & 64.3 & 54.55 & 71.05 & 63.24 & 73.73 & 68.08 \\
    \hline WA \textit{w/} Neural Compressor & \textbf{62.76} & \textbf{52.81} & \textbf{63.84} & \textbf{56.18} & \textbf{73.3} & \textbf{66.56} & \textbf{75.64} & \textbf{71.4}\\ \hline
    \hline PODNet~\cite{PODNet} \textit{w/o} compression & 50.03 & 29.19 & 54.97 & 35.96 & 54.88 & 35.9 & 58.98 & 40.94 \\
    \hline PODNet \textit{w/} JPEG & 45.86 & 31.37 & 47.21 & 33.62 & 65.96 & 55.98 & 67.89 & 58.8 \\
    \hline PODNet \textit{w/} WebP & 55.03 & 40.16 & 55.29 & 41.64 & 65.23 & 53.64 & 67.82 & \textbf{59.16} \\
    \hline PODNet \textit{w/} Neural Compressor & \textbf{56.36} & \textbf{42.79} & \textbf{56.15} & \textbf{42.58} & \textbf{68.61} & \textbf{61.5} & \textbf{68.01} & 58.96\\ \hline
    \hline FOSTER~\cite{Foster} \textit{w/o} compression & 61.75 & 45.63 & 64.32 & 52.3 & 68.88 & 57.46 & 69.60 & 61.52 \\
    \hline FOSTER \textit{w/} JPEG & 52.29 & 37.51 & 55.6 & 43.67 & 74.82 & 68.24 & 76.99 & 71.86 \\
    \hline FOSTER \textit{w/} WebP & 64.92 & 52.17 & 67.26 & 57.74 & 73.4 & 65.56 & 76.16 & 70.46 \\
    \hline FOSTER \textit{w/} Neural Compressor & \textbf{66.81} & \textbf{55.87} & \textbf{68.18} & \textbf{60.01} & \textbf{75.53} & \textbf{69.3} & \textbf{77.73} & \textbf{72.98}\\
    \hline
    \end{tabular}
\caption{Accuracy (\%) of several baseline CIL methods with and without compression on two datasets using the learning from scratch (LFS) protocol. Each CIL algorithm with compression methods determine the compression rate using our proposed method. Scale Hyperprior~\cite{minnen2018joint} with quality=1 for ImageNet-100 and Factorized model~\cite{balle18hyperprior} with quality=5 for CIFAR-100 are the final selected algorithms.}\vspace{-0.3cm}
\label{tab:Acc_Tab}
\end{table*}

\subsection{Dataset}
In adherence to the protocol of class incremental learning outlined in~\cite{iCaRL}, two public datasets are used for evaluation: CIFAR-100~\cite{Cifar100}, which is uncompressed, and ImageNet-100~\cite{ImageNet}, which is a JPEG compressed. CIFAR-100 contains 100 classes with 50k images for training and 10k for testing with the size of 32$\times$32. ImageNet-100 is a subset of the large-scale dataset ImageNet-1000~\cite{ImageNet}, which contains around 1.2 million training images and 50k validation images, with varying image sizes. It is important to note that ImageNet-100 is selected from the first 100 classes after a random shuffle. Consistent with the protocol in~\cite{iCaRL}, all classes are shuffled using the random seed of 1993. These classes are then equally partitioned into 10 tasks, where each task contains 10 classes in accordance with the Learn From Scratch (LFS) setting following recent CIL methods~\cite{Foster, CIM_CIL}. 


\subsection{Experiment Setup}
For fair comparison, all methods employ ResNet32~\cite{resnet} for CIFAR-100 and ResNet18 for ImageNet-100 as backbone network, with a simple fully-connected layer serving as the classifier following the standard setup in~\cite{iCaRL}. We also kept the hyperparameters consistent across all methods. Stochastic Gradient Descent (SGD) optimizer is utilized with an initial learning rate of 0.1 and momentum of 0.9. The training process is set for 200 training epochs for the initial phase and 170 for subsequent phases. The learning rate performs a decay of 0.1 at 80 and 120 epochs. The batch size is set to 64, and a consistent random seed, 1993, is used for all experiments. 

We report several representative CIL methods with memory replay, including iCaRL~\cite{iCaRL}, WA~\cite{wa2020}, PODNet~\cite{PODNet} and FOSTER~\cite{Foster}, and investigate the effect of compressed exemplars using three different compression method, namely JPEG, WebP and Neural Compressors~\cite{balle18hyperprior,minnen2018joint}. Note that the neural compressors are pre-trained on the Vimeo90k dataset~\cite{xue2019video} instead of CIFAR-100 or ImageNet-100 for two reasons. First, the full dataset of CIFAR-100 and ImageNet-100 are not available during training as data comes in sequence under CIL scenarios. Second, employing pre-trained image compression methods in vision tasks such as classification is a common practice in compression literature~\cite{10029924,rebuttal2}. This approach is justified for two key reasons. First, compression models are semantically agnostic, as they are trained without the use of labels. Second, the training datasets used for compression tasks (e.g., Vimeo90K) are distinct from those employed in CIL contexts (e.g., CIFAR-100 and ImageNet-100). We therefore treat these compression algorithms as pre-processing for CIL model training and testing.

\vspace{-0.24cm}
\subsection{Results and Analysis}
Results are reported in Table~\ref{tab:Acc_Tab}. For each method, we report the baseline results without compression, and the results with compression using JPEG, WebP, and Neural Compressor. Note that the rates are selected using our method in Section~\ref{subsec:suitableRes}. we report the top-1 average accuracy of all steps to illustrate the model's overall performance throughout the entire continual learning procedure, and the final step's top-1 accuracy to demonstrate the performance after finishing the entire continual learning procedure for all seen classes. Two different buffer sizes are evaluated for each dataset, 2.93MB and 5.86 MB for CIFAR-100, and is 143.5MB and 287MB for ImageNet-100. The buffer sizes correspond to the equivalent of 1,000 and 2,000 original images for each dataset.

In Table~\ref{tab:Acc_Tab}, we evaluate various methods with and without compression on the uncompressed CIFAR-100 dataset, which was originally formatted as integer arrays, and JPEG-compressed ImageNet-100 dataset. The compression rate \& algorithm we selected in Section~\ref{subsec:selection}, which is Factorized model with quality=5 for CIFAR-100 and Scale Hyperprior model with quality=1 for ImageNet-100, generally obtains the highest accuracy among the three compression algorithms in all four CIL methods. 
Across all four CIL methods, neural compressor and WebP show better performance compared with JPEG on CIFAR-100, which is consistent with our results using $F_{MSE}$ in Table~\ref{tab:metric_cifar100}. For ImageNet-100, we observe that across all four CIL methods, neural compression and JPEG show better performance compared with WebP, which seems inconsistent with our results on $F_{MSE}$ in Table~\ref{tab:metric_imageNet100}. According to Table~\ref{tab:metric_imageNet100}, while JPEG exhibits the largest $F_{MSE}$, the final accuracy is not the worst among the three compression methods for all four CIL methods. This could be attributed to the inherent characteristics of the dataset, as ImageNet-100 is initially JPEG-compressed. Although additional JPEG compression results in greater spatial and feature domain distortion, the model seems to adapt well, compared to using other compression methods.

From Table~\ref{tab:Acc_Tab}, we can make the following observations. First, JPEG does not work well for small images (32$\times$32) compared to neural compressors and WebP, hence showing lower accuracy on CIFAR-100 as shown in Table \ref{tab:Acc_Tab}. Second, when properly selecting compression algorithm and rate, it significantly enhances both the average and last-step performance compared to uncompressed methods, which is consistent across all tested methods. Third, the improvement in average accuracy is particularly evident under strict constraints on memory settings, such as smaller buffer sizes.

\vspace{-0.3cm}
\subsection{Comparison with Other Compression-based Methods}
Table~\ref{tab:Acc_Tab_comparison} presents a comparison of our results with two other compression-based methods: MRDC~\cite{Wang2022MemoryRW} and CIM~\cite{CIM_CIL}. CIM~\cite{CIM_CIL} downsampled only non-discriminative pixels, and MRDC~\cite{Wang2022MemoryRW} utilize relatively high JPEG rates to compress data to ensure exemplar quality. However, our method successfully expands the diversity of exemplars by adopting low data rate compression, without compromising on the quality of the exemplars. For a more comprehensive comparison, we included not only LFS but also the Learn from Half (LFH) protocol, where the classes are separated into six tasks, with the first task containing 50 classes and each of the rest containing 10 classes. Our method consistently surpasses both existing methods in the LFS and LFH settings. This improvement primarily stems from the utilization of low data rate compression in our method, coupled with a strategy to counteract domain shift by pre-processing data through compression. This makes the model adapt to compression artifacts and also constructs a more balanced and diverse dataset throughout each incremental training step. The strength of our method shows more when more incremental tasks are involved (10 for LFS compared with 6 for LFH), as our method is able to handle the forgetting issue better in CIL setups. 
\begin{table}[h]
    \centering
    \begin{tabular}{|c|c|c|c|c|c|c|c|c|c|c|c|c|}
    \hline \multirow{3}{*}{$\begin{array}{c}\text {Dataset} \\
    \text {Protocol} \\
    \text { Accuracy (\%) } \\
    \end{array}$} & \multicolumn{4}{|c|}{ImageNet-100} \\
    \hline & \multicolumn{2}{|c|}{LFS} & \multicolumn{2}{|c|}{LFH} \\
    \hline & Avg & Last & Avg & Last\\
    \hline FOSTER~\cite{Foster} &74.73  &67.26  &  73.96&71.0  \\
    \hline \textit{w/} MRDC~\cite{Wang2022MemoryRW}&  78.13&  73.36&  73.26& 74.16 \\
    \hline \textit{w/} CIM \cite{CIM_CIL}&  77.27&  67.64& \textbf{79.18}&  72.12 \\
    \hline \textit{w/} Ours & \textbf{79.40} & \textbf{75.52} & 74.90 & \textbf{75.98}\\
    \hline
    \end{tabular}
\caption{Comparison with existing CIL methods using compressed exemplars including MRDC~\cite{Wang2022MemoryRW} and CIM~\cite{CIM_CIL}. We plug each of them in baseline FOSTER and follow the training settings of CIM for fair comparison. The memory buffer size is 287MB.}
\label{tab:Acc_Tab_comparison}
\end{table}
\vspace{-0.4cm}
\vspace{-0.2cm}
\subsection{Discussion}
While rate-distortion (RD) curves have shown to be effective for evaluating compression models in the context of image reconstruction quality, a good RD curve, shown in Figure~\ref{fig:dataset_rd}, does not inherently translate to enhanced classification performance in CIL settings, as shown in Table~\ref{tab:metric_cifar100} and Table~\ref{tab:metric_imageNet100}. The RD curve is mainly used in the field of data compression. Although a superior RD curve demonstrates better compression performance, it does not directly signify an improved classification accuracy for CIL models. The effectiveness of CIL models relies on several key factors, such as resilience to catastrophic forgetting, the ability to generalize from old tasks to new ones, and its robustness when facing dataset shifts across various tasks. Consequently, although the improved RD curve can bolster data efficiency and storage, it is not able to serve as a sufficient measure to determine the effectiveness of model classification performance under CIL scenarios. Hence we propose to use the forgetting measure for rate selection, and $F_{MSE}$ for compression algorithm selection, as stated in Section~\ref{subsec:selection} that aligns better with the final accuracy.

\vspace{-0.3cm}
\section{Conclusion}
\label{sec:conclusion}
\vspace{-0.1cm}
Real-world ML systems need to learn from sequential data rather than a static dataset for model training. The ability for models to continually update without irreversible catastrophic forgetting is crucial for practical applications. In this paper, we study the effect of incorporating compression algorithms into memory replay-based CIL methods. We establish a benchmark to conduct fair comparisons across different memory settings for CIL scenarios. Furthermore, we propose an efficient strategy for selecting a proper compression rate and method to attain a good balance between quantity and quality. By harnessing compression methods, we design a more resource-efficient approach to build continual ML systems without compromising performance under limited memory constraints.

{
\small
\bibliographystyle{unsrt}
\bibliography{main}

\begin{thebibliography}{10}

\bibitem{PODNet}
A.~Douillard, M.~Cord, C.~Ollion, T.~Robert, and E.~Valle.
\newblock Podnet: Pooled outputs distillation for small-tasks incremental learning.
\newblock {\em European Conference on Computer Vision}, page 86–102, August 2020.

\bibitem{iCaRL}
S.~A. Rebuffi, A.~Kolesnikov, G.~Sperl, and C.~H. Lampert.
\newblock icarl: Incremental classifier and representation learning.
\newblock {\em IEEE/CVF Conference on Computer Vision and Pattern Recognition}, June 2017.

\bibitem{mnemonics}
Y.~Liu, Y.~Su, A.~A. Liu, B.~Schiele, and Q.~Sun.
\newblock Mnemonics training: Multi-class incremental learning without forgetting.
\newblock {\em IEEE/CVF Conference on Computer Vision and Pattern Recognition}, 2020.

\bibitem{Liu2021RMM}
Y.~Liu, B.~Schiele, and Q.~Sun.
\newblock {RMM:} reinforced memory management for class-incremental learning.
\newblock {\em Advances in Neural Information Processing Systems}, pages 3478--3490, December 2021.

\bibitem{janeiro2023visual}
J.~M. Janeiro, S.Frolov, A.~El-Nouby, and J.~Verbeek.
\newblock Are visual recognition models robust to image compression?
\newblock {\em Neural Compression Workshop, International Conference on Machine Learning}, 2023.

\bibitem{wallace1992jpeg}
G.~Wallace.
\newblock The jpeg still picture compression standard.
\newblock {\em IEEE Transactions on Consumer Electronics}, 38(1):xviii--xxxiv, February 1992.

\bibitem{ding2021advances}
D.~Ding, Z.~Ma, D.~Chen, Qingshuang Q.~Chen, Z.~Liu, and F.~Zhu.
\newblock Advances in video compression system using deep neural network: A review and case studies.
\newblock {\em Proceedings of the IEEE}, 109(9):1494--1520, 2021.

\bibitem{Chaudhry_2018_ECCV}
A.~Chaudhry, P.~K. Dokania, T.~Ajanthan, and P.~H.~S. Torr.
\newblock Riemannian walk for incremental learning: Understanding forgetting and intransigence.
\newblock {\em European Conference on Computer Vision}, September 2018.

\bibitem{herding}
M.~Welling.
\newblock Herding dynamical weights to learn.
\newblock {\em International Conference on Machine Learning}, page 1121–1128, 2009.

\bibitem{CIM_CIL}
Z.~Lou, Y.~Liu, B.~Schiele, and Q.~Sun.
\newblock Class-incremental exemplar compression for class-incremental learning.
\newblock {\em IEEE Conference on Computer Vision and Pattern Recognition}, June 2023.

\bibitem{Wang2022MemoryRW}
L.~Wang, X.~Zhang, K.~Yang, L.~L. Yu, C.~Li, L.~Hong, S.~Zhang, Z.~Li, Y.~Zhong, and J.~Zhu.
\newblock Memory replay with data compression for continual learning.
\newblock {\em International Conference on Learning Representations}, April 2022.

\bibitem{Foster}
F.~Wang, D.~Zhou, H.~Ye, and D.~Zhan.
\newblock Foster: Feature boosting and compression for class-incremental learning.
\newblock {\em European Conference on Computer Vision}, October 2022.

\bibitem{balle18hyperprior}
J.~Ballé, D.~Minnen, S.~Singh, S.~Hwang, and N.~Johnston.
\newblock Variational image compression with a scale hyperprior.
\newblock {\em International Conference on Learning Representations}, April 2018.

\bibitem{minnen2018joint}
D.~Minnen, J.~Ball\'{e}, and G.~Toderici.
\newblock Joint autoregressive and hierarchical priors for learned image compression.
\newblock {\em Advances in Neural Information Processing Systems}, 31:10794--10803, December 2018.

\bibitem{resnet}
K.~He, X.~Zhang, S.~Ren, and J.~Sun.
\newblock Deep residual learning for image recognition.
\newblock {\em IEEE/CVF Conference on Computer Vision and Pattern Recognition}, pages 770--778, 2016.

\bibitem{wa2020}
B.~Zhao, X.~Xiao, G.~Gan, B.~Zhang, and S.~Xia.
\newblock Maintaining discrimination and fairness in class incremental learning.
\newblock {\em IEEE/CVF conference on computer vision and pattern recognition}, June 2020.

\bibitem{Cifar100}
A.~Krizhevsky and G.~Hinton \textit{et al}.
\newblock Learning multiple layers of features from tiny image.
\newblock {\em Technical Report}, 2009.

\bibitem{ImageNet}
J.~Deng, W.~Dong, R.~Socher, L.~J. Li, L.~Kai, and F.~F. Li.
\newblock Imagenet: A large-scale hierarchical image database.
\newblock {\em IEEE Conference on Computer Vision and Pattern Recognition}, pages 248--255, June 2009.

\bibitem{xue2019video}
T.~Xue, B.~Chen, J.~Wu, D.~Wei, and W.~T. Freeman.
\newblock Video enhancement with task-oriented flow.
\newblock {\em International Journal of Computer Vision}, 2019.

\bibitem{10029924}
Z.~Duan, Z.~Ma, and F.~Zhu.
\newblock Unified architecture adaptation for compressed domain semantic inference.
\newblock {\em IEEE Transactions on Circuits and Systems for Video Technology}, 33(8):4108--4121, 2023.

\bibitem{rebuttal2}
R.~Torfason, F.~Mentzer, E.~Agustsson, M.~Tschannen, R.~Timofte, and L.~V. Gool.
\newblock Towards image understanding from deep compression without decoding.
\newblock {\em International Conference on Learning Representations}, 2018.

\end{thebibliography}
}

\end{document}